\documentclass[letterpaper]{article} 
\usepackage{aaai2027}  
\usepackage{helvet}
\usepackage{cou rier}
\usepackage[hyphens]{url}  
\usepackage{graphicx} 
\usepackage{natbib}  
\usepackage{caption} 
\usepackage{amsfonts} 
\usepackage{amsmath} 
\usepackage{algorithm}
\usepackage{algorithmic}
\usepackage{cleveref}
\usepackage{multirow}
\crefname{section}{Sect.}{Sects.}
\crefname{figure}{Fig.}{Figs.}
\crefname{table}{Tab.}{Tabs.}
\crefname{equation}{Eq.}{Eqs.}

\usepackage{newfloat}
\usepackage{listings}
\DeclareCaptionStyle{ruled}{labelfont=normalfont,labelsep=colon,strut=off} 
\floatstyle{ruled}
\newfloat{listing}{tb}{lst}{}
\floatname{listing}{Listing}

\usepackage{booktabs}

\newcommand{\ieno}{\textit{i.e.}}

\title{MRBench: A Comprehensive Benchmark for Human Motion-Text Retrieval}

\author{
Fulong Liu\textsuperscript{\rm 1,*},
Liang Xu\textsuperscript{\rm 2,*},
Chengqun Yang\textsuperscript{\rm 1},
Yuhao Zhang\textsuperscript{\rm 1},
Yichao Yan\textsuperscript{\rm 1,$\dagger$},
Xiaokang Yang\textsuperscript{\rm 1}
}

\affiliations{
\textsuperscript{\rm 1}Shanghai Jiao Tong University \quad
\textsuperscript{\rm 2}Beijing Zhongguancun Academy\\
\textsuperscript{*}Equal contribution. \quad
\textsuperscript{$\dagger$}Corresponding author.
}

\usepackage{bibentry}
\begin{document}

\maketitle

\begin{abstract}
Human motion-text retrieval provides a rigorous means of assessing cross-modal alignment.
Prevailing benchmarks are dominated by homogeneous indoor motions, imbalanced motion distributions, and oversimplified, repetitive texts, which hinder the reliable measurement of cross-domain and cross-granularity alignment.
We thus introduce MRBench, a comprehensive motion-text retrieval benchmark featuring heterogeneous motions, broad and balanced category coverage, and reliable, discriminative, multi-granular descriptions. 
MRBench is constructed through a meticulously designed multi-stage data curation pipeline, which filters and balances candidates, verifies unambiguous semantic alignment, and generates motion-grounded descriptions at multiple granularities.
The resulting benchmark contains 3,390 motions drawn from motion capture, in-the-wild videos, synthetic videos, and motion generative models, covering 118 fine-grained categories. 
Each motion is paired with concise, standard, and fine-grained descriptions, yielding 10,170 captions.
Extensive evaluations of representative retrieval baselines on MRBench reveal a substantial cross-dataset generalization gap and pronounced sensitivity to query granularity. 
We propose a lightweight granularity-aware model anchored at a frozen standard-caption-aligned retrieval model. LLM-based concise and fine-grained captions provide pseudo-supervision for extra-branch granularity-specific motion extractors and text adapters.
For inference, granularity-aware score fusion integrates global and adapted similarities while strictly maintaining score comparability across all description levels.
The resulting model improves mixed-granularity retrieval without compromising standard-caption performance.
We believe that our MRBench provides a comprehensive testbed for advancing motion-language alignment evaluation.
\end{abstract}

\vspace{-3pt}
\section{Introduction}
\label{sec:intro}
The understanding and generation of human motion are pivotal and highly challenging in computer vision~\cite{kinmo2025kinematicawarehumanmotion,  zhuMotionGPT3HumanMotion2025, guoMoMaskGenerativeMasked2023a} and robotics~\cite{luo2025sonic,jiang2025uniact,shao2025langwbc}. Despite decades of continuous dataset scaling and algorithmic evolution~\cite{guo2020action2motion, punnakkalBABELBodiesAction2021, hwang2026snapmogen, fanGoZeroMotionMillion2025, cao2025being}, the unified, open-world motion foundation models~\cite{li2025genmo, lu2025scamo, cao2025being, wang2026motionbricks} remain largely elusive, where one of the fundamental challenges is to establish a robust and powerful alignment between motion and text.

In this paper, we primarily focus on human motion-text retrieval~\cite{petrovichTMRTexttoMotionRetrieval2023,yu2024exploring,zhangsgar2024} as a direct and rigorous proxy for evaluating this cross-modal alignment capability, where existing efforts still suffer from several limitations for both modalities. From the perspective of motion, human behaviors exhibit long-tailed distributions, which may lead to extreme data bias and imbalance in motion datasets. Furthermore, 3D motions are inherently subtle with rich kinematic details~\cite{SMPL-X:2019, zhou2019continuity}. Conversely, from a linguistic perspective, text descriptions naturally span a wide range of granularities, from condensed summaries to fine-grained depictions. We emphasize that a comprehensive benchmark should simultaneously accommodate the characteristics of human motion together with the ambiguity of multi-granular textual descriptions.

Revisiting existing motion-text retrieval benchmarks, prior works predominantly rely on HumanML3D~\cite{humanml3d} and KIT-ML~\cite{plappertKITMotionLanguage2016} for evaluation. We argue and empirically verify that these benchmarks exhibit critical limitations in faithfully reflecting cross-modal alignment capability, which significantly hinders progress in this domain. The motion data is homogeneous based on indoor motion capture (MoCap) scenarios, which is narrowly distributed and extremely imbalanced (\ieno, dominated by ``walking'' in HumanML3D). Besides, the textual descriptions are overly simple and suffer from severe repetition. For instance, a single text prompt could be mapped to hundreds of distinct motion instances. Consequently, these systemic flaws invalidate standard retrieval metrics, underscoring the urgent need for a comprehensive and reliable benchmark paradigm.

\begin{figure*}[t]
    \centering
    \includegraphics[width=\textwidth]{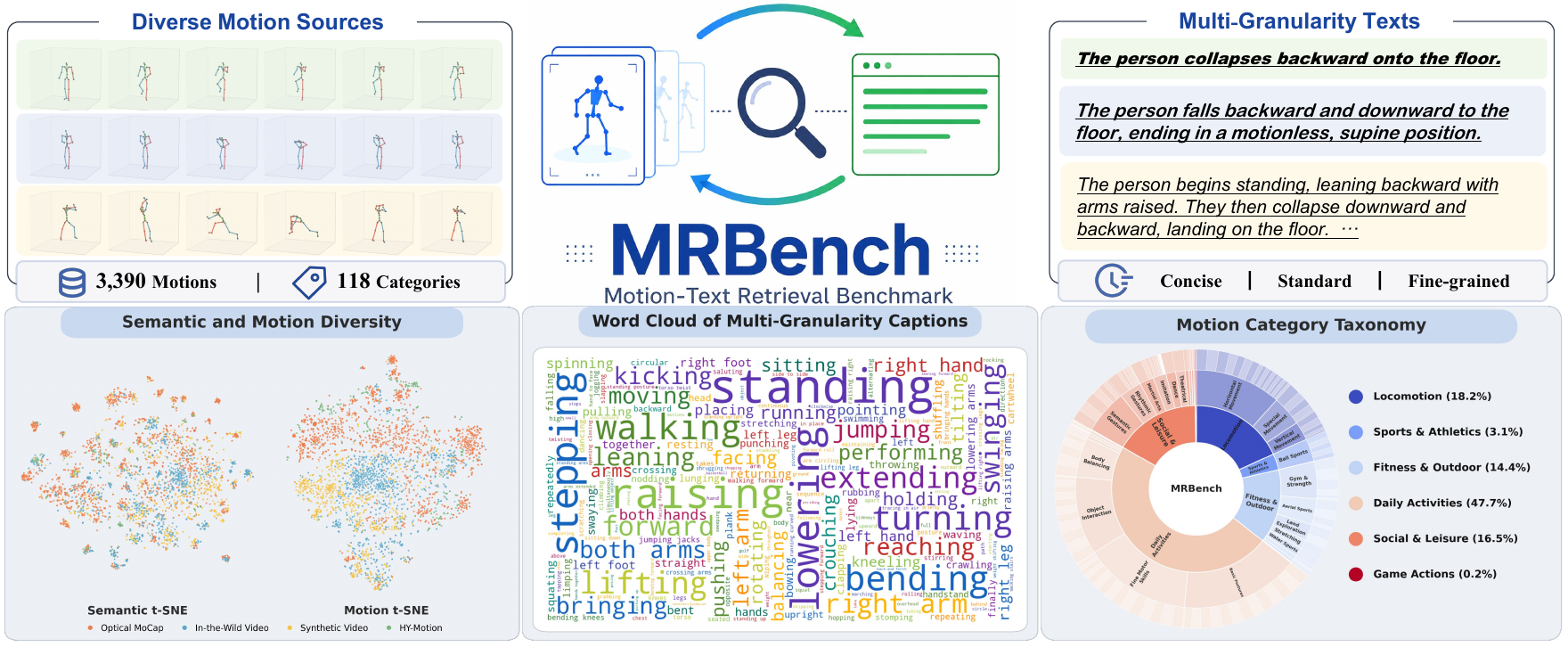}
    \setlength{\abovecaptionskip}{-8pt}
    \caption{
    \textbf{Overview of MRBench.} MRBench features diverse motion sources, broad motion-category coverage, and multi-granularity text descriptions, enabling a more comprehensive evaluation of human motion-text retrieval.
    }
    \label{fig:teaser}
\end{figure*}

Motivated by these meticulous analytical insights, we introduce \textbf{MRBench} (shown in~\cref{fig:teaser}), the first large-scale, multi-source benchmark specifically tailored for comprehensive human motion-text retrieval evaluation. 
To fundamentally overcome the homogeneous nature and distributional bias of prior datasets, MRBench comprises a multi-source, more balanced motion collection that drastically expands the domain space by integrating indoor Mocap data with in-the-wild video-derived motions~\cite{vimogen}, synthetic video-derived motions, and generated motions~\cite{hymotion}. 
Based on the ViMoGen-228K~\cite{vimogen} dataset, we first remove generic, repetitive, and non-kinematic captions through rule-based filtering and LLM-based semantic scoring, and then perform taxonomy-guided balanced sampling to enforce categorical coverage and remove semantically redundant captions.
Furthermore, semantic alignment assessment is performed by large vision-language models to distill well-aligned and unambiguous motion-text pairs.
Finally, visually grounded rewriting produces concise, standard, and fine-grained descriptions with hierarchical granularity, followed by manual validation, to facilitate cross-granularity motion-text alignment evaluation.
The resulting benchmark contains \textbf{3,390} motion sequences and covers \textbf{118} fine-grained motion categories. Each motion is annotated at three textual granularities, yielding \textbf{10,170} motion-verified captions.


We extensively evaluate existing retrieval models across heterogeneous motion sources and three levels of textual granularity on MRBench, revealing substantial cross-dataset generalization gaps and pronounced sensitivity to query granularity. To mitigate this issue, we introduce a lightweight, granularity-aware retrieval model. 
We first train a vanilla retrieval model on standard-caption pairs and freeze its encoders and global projection heads as an alignment anchor. Concise and fine-grained captions rewritten by LLMs then serve as pseudo-supervision for learning granularity-specific attention-based motion extractors and text projection adapters. For inference, standard descriptions use the frozen global branch, whereas non-standard descriptions combine global and adapted similarities through granularity-aware score fusion. This design improves fine-grained and mixed-granularity retrieval while preserving standard-caption performance.
Our main contributions can be summarized as:
\begin{itemize}
    \item We construct MRBench as the first comprehensive human motion-text retrieval benchmark that features multi-source motion distributions and natively supports cross-granularity motion-text alignment evaluation.
    \item We propose a multi-stage data curation pipeline that combines rule-based filtering, taxonomy-guided balanced sampling, cross-modal semantic assessment, and multi-granular rewriting with human validation.
    \item We propose a lightweight granularity-aware model that freezes a standard-caption-aligned dual encoder and uses LLM-rewritten captions to supervise granularity-specific motion extractors and text adapters, preserving standard-caption alignment while supporting non-standard queries.
\end{itemize}

\begin{figure*}[t]
    \centering
    \includegraphics[width=\textwidth]{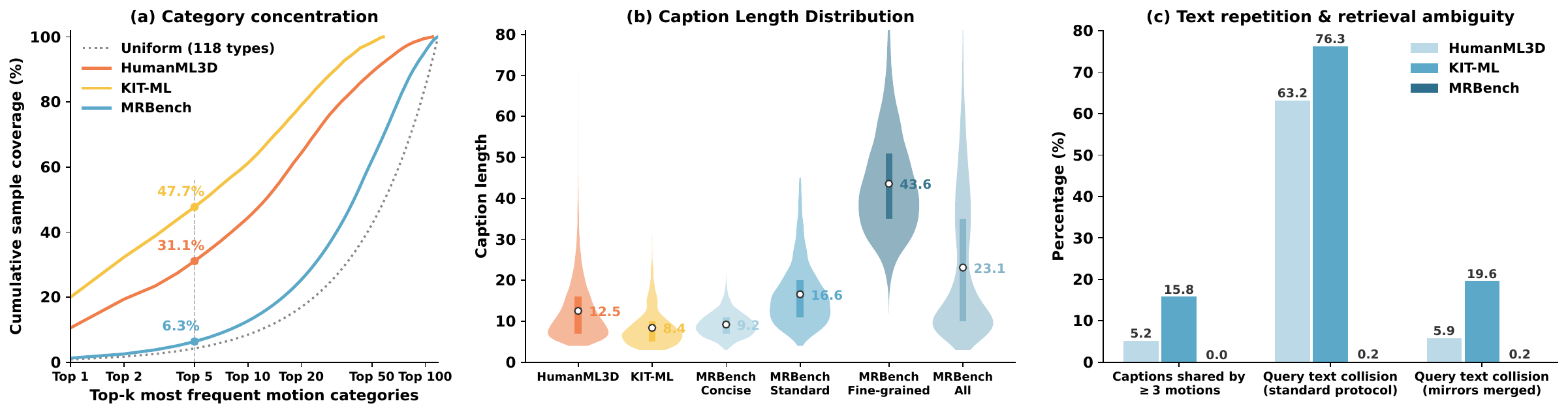}
    \setlength{\abovecaptionskip}{-8pt}
    \caption{Statistical comparisons between MRBench and existing benchmarks, computed on the official test splits. \textbf{(a)}~Cumulative sample coverage of the top-$k$ most frequent motion categories.
    \textbf{(b)}~Caption length distributions (white dots denote medians); MRBench provides three granularities of Concise, Standard, and Fine-grained per motion. 
    \textbf{(c)}~Text repetition and the resulting evaluation ambiguity: the proportion of captions shared verbatim by $\geq$3 motions, and the proportion of test queries whose text also describes at least one other gallery motion under the standard protocol and after merging mirrored pairs.}
    \label{fig:benchmark_bias}
\end{figure*}

\vspace{-3pt}
\section{Revisiting Motion-Text Retrieval}
Current human motion-text retrieval is predominantly evaluated on HumanML3D~\cite{humanml3d} and KIT-ML~\cite{plappertKITMotionLanguage2016}, both of which are curated for motion generation yet problematic for retrieval.
\textbf{Firstly}, the motion modality suffers from severe homogeneity in data provenance and category distribution. The indoor MoCap collections fail to probe the kinematic diversity of in-the-wild motions derived from videos or generative models. As shown in~\cref{fig:benchmark_bias}(a), the category distribution is heavily concentrated, where the top-5 most frequent motion types already account for 47.7\% for KIT-ML and 31.1\% for HumanML3D, and KIT-ML covers only 58 fine-grained motion types in total. 
\textbf{Secondly}, the text annotations are oversimplified and highly redundant. The median caption length is 7 words for KIT-ML and 11 for HumanML3D as depicted in~\cref{fig:benchmark_bias}(b). Besides, 15.8\% / 5.2\% of KIT-ML / HumanML3D test captions are shared verbatim by at least three different motions in~\cref{fig:benchmark_bias}(c). Most critically, generic texts such as \textit{``a person slowly walked forward''} are attached to as many as 148 distinct HumanML3D motions, which is extremely detrimental to retrieval evaluation.
Under the standard single-positive protocol, 76.3\% of KIT-ML and 63.2\% of HumanML3D test queries share identical text with at least one other gallery motion, thus a model retrieving an equally valid motion is nevertheless counted as wrong; even after merging trivial mirror-augmented pairs, 19.6\% and 5.9\% of queries remain intrinsically ambiguous. Consequently, reporting Recall@$k$ under such extreme ambiguity conflates genuine model capability with annotation artifacts, failing to provide a reliable indicator for fine-grained language understanding. 
These observations motivate a benchmark with heterogeneous motion sources, balanced category coverage, and unique, multi-granular, motion-verified, and well-aligned descriptions, which we introduce next.

\vspace{-2pt}
\section{MRBench for Motion-Text Retrieval}
\subsection{Benchmark Construction}
\begin{figure*}[t]
    \centering
    \includegraphics[width=\textwidth]{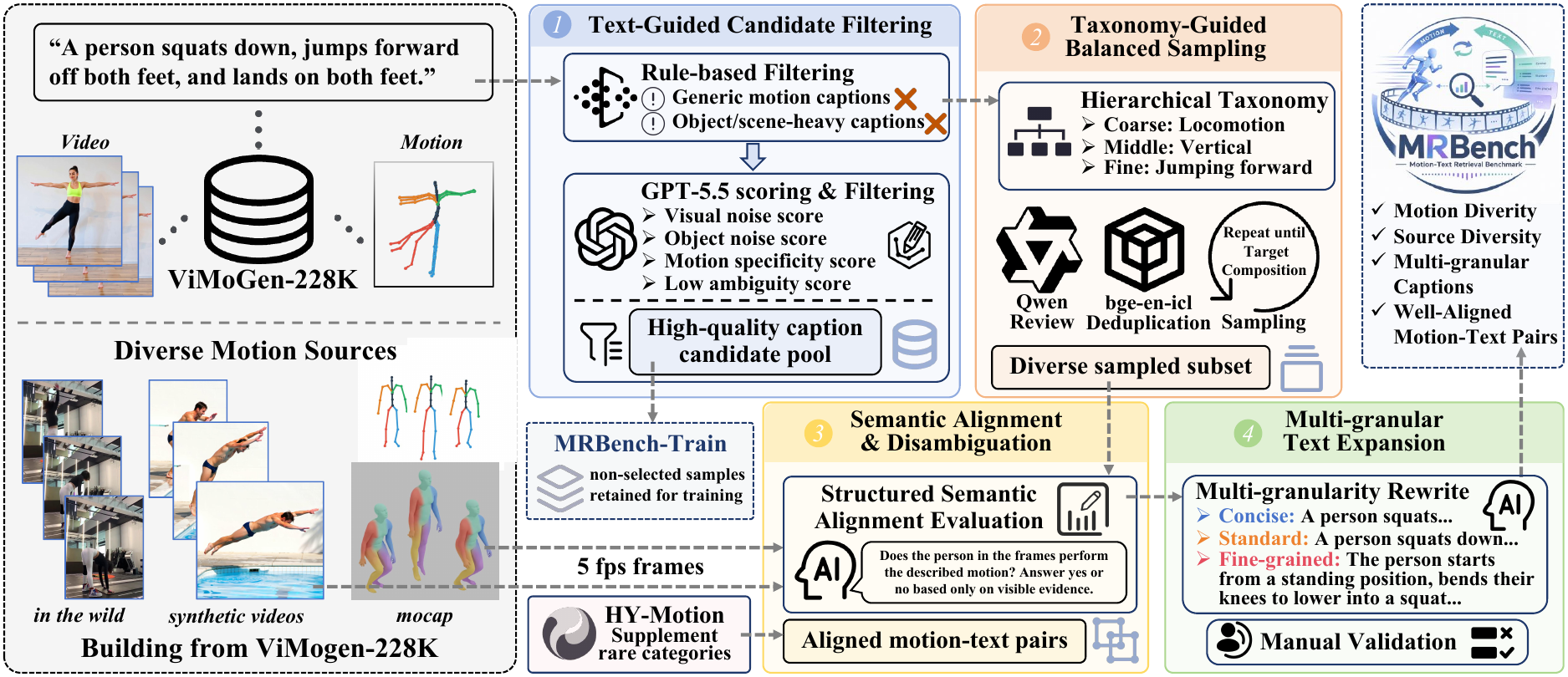}
    \setlength{\abovecaptionskip}{-8pt}
    \caption{Overview of the MRBench construction pipeline. Starting from ViMoGen-228K, we perform retrieval-oriented caption filtering, hierarchical taxonomy-based sampling, motion-text alignment verification, and targeted supplementation, followed by multi-granular annotation and human validation.}
    \label{fig:pipeline}
    \vspace{-2pt}
\end{figure*}

\vspace{-2pt}
To construct a comprehensive human motion-text retrieval benchmark, we develop a rigorous, multi-dimensional curation pipeline based on the ViMoGen-228K~\cite{vimogen} corpus, which offers rich motion diversity and semantic coverage. Our pipeline is designed to retain rich diversity while providing reliable, unambiguous, and multi-granular motion-text alignments.
The overall construction procedure consists of four major stages, as detailed in~\cref{fig:pipeline}.

\noindent \textbf{Text-Guided Candidate Filtering.}
Given the large-scale motion-text pool, we initially filter the data from the textual perspective for its explicit semantic cues. We notice that indoor-MoCap motions often contain highly generic and repetitive expressions, such as \textit{``A person walks forward''}, which provide limited semantic information for distinguishing visually similar motions. For video-derived motion captions, such as \textit{``As she continues her walk, the garden's beauty unfolds around her, offering a peaceful escape from the world''}, excessive non-kinematic details such as object, scene, appearance, and affection severely distract the actual alignment.
We therefore first employ rule-based filtering to remove identical repetitive and overly generic expressions. 
The remaining captions are subsequently assessed by GPT-5.5~\cite{openai2026gpt55} to score their semantic discriminability and compactness. We then filter out redundant yet non-kinematic textual descriptions as evaluation candidates. The high-quality samples that are not selected for the final benchmark are retained as the primary training set in our subsequent experiments, referred to as \textbf{MRBench-Train}.

\noindent \textbf{Taxonomy-Guided Balanced Sampling.}
To obtain a more balanced and diverse evaluation set, we adopt a taxonomy-guided balanced sampling strategy. We first organize the candidates into a coarse-to-fine taxonomy as shown in~\cref{fig:teaser} inspired by the hierarchical categorization of HY-Motion~\cite{hymotion}. Samples with explicit motion semantics like \textit{``walking forward''} can be directly assigned to the \textit{``Locomotion/Horizontal''} category by rules, while remaining cases that cannot be reliably categorized are further classified by Qwen2.5-7B-Instruct~\cite{qwen2025qwen25technicalreport}.
The taxonomy organization facilitates a category-aware structured sampling mechanism that ensures comprehensive coverage across diverse motion categories. For densely populated categories such as \textit{``Daily Activities''}, we leverage bge-en-icl~\cite{li2024makingtextembeddersfewshot} embeddings to remove semantically redundant captions and regulate the balance and uniformity.

\noindent \textbf{Semantic Alignment and Disambiguation.}
The preceding two stages perform filtering within the textual modality but cannot guarantee precise and unambiguous motion-text alignment. We therefore adopt Structured Semantic Alignment Evaluation (SSAE), following HY-Motion~\cite{hymotion}, to further filter candidate motion-text pairs based on their cross-modal consistency. We leverage the powerful vision-language alignment capabilities of advanced multimodal models to assess the correspondence between texts and motion frames, \ieno, extracted from rendered, in-the-wild, and synthetic motion videos. Specifically, we uniformly sampled at 5\,FPS and evaluated by Gemini-3-pro-preview~\cite{googledeepmind2025gemini3pro}, which determines whether the primary motion described by the caption is observable in the corresponding frames and provides an explicit rationale for its decision.
We also perform rolling sampling from the second-stage taxonomy pool to compromise the dataset's structural balance. 
For severely underrepresented motion categories, we adopt HY-Motion to synthesize motions based on GPT-5.5-generated motion scripts. This selection, filtering, and supplementation is repeated until the balanced, well-aligned, and unambiguous benchmark composition is reached.

\noindent \textbf{Multi-granular Text Expansion.}
We enrich the benchmark with multi-granular textual annotations to measure the hierarchical cross-modal correspondence between motion and text. The motion, text, and alignment rationale are provided jointly as context for Gemini-3-pro-preview to generate concise, fine-grained descriptions that respectively capture the condensed principal semantics and detailed body-part and temporal dynamics. Unlike text-only paraphrasing, this visually grounded rewriting process ensures that descriptions at different granularities remain consistent with the observed motion.
After rewriting, we manually review the results to check the overall annotation quality, taxonomy assignments, and validity of hierarchical texts.

Notably, we employ different large models across all stages to reduce annotation biases and avoid particular linguistic styles. Distributing scoring, categorization, alignment verification, and text expansion across multiple models helps preserve textual diversity and reduces model-specific bias, while maintaining the semantic precision required for a challenging motion-text retrieval benchmark.

\subsection{Benchmark Statistics}
\vspace{-2pt}
MRBench contains \textbf{3,390} motion sequences, each paired with descriptions at three granularities, yielding \textbf{10,170} captions in total. All motions are standardized to a unified skeletal representation at 20\,FPS, with durations ranging from 1.6\,s to 17.5\,s and averaging 6.2\,s, totaling approximately 5.8 hours. MRBench exhibits three key characteristics:

\begin{itemize}
\item \textbf{Multi-source composition:} MRBench integrates 1,484 MoCap sequences, 899 in-the-wild video-derived motions, 770 synthetic video-derived motions, and 237 HY-Motion-generated sequences, which account for 43.8\%, 26.5\%, 22.7\%, and 7.0\% of the benchmark, respectively. This heterogeneous composition mitigates the motion distribution biases of indoor MoCap-only benchmarks.

\item \textbf{Balanced semantic coverage:} All samples are organized under a hierarchical taxonomy with 7 coarse, 33 middle-level, and \textbf{118} fine-grained motion categories. Through stratified sampling and targeted supplementation, each motion type contains 10$\sim$60 samples with a median of 42, avoiding the severe long-tailed distribution.

\item \textbf{Distinct, multi-granular texts:} All descriptions are visually grounded through SSAE-based verification and rewriting; the resulting motion-text pairs are highly discriminative and nearly duplicate-free.
Each motion is annotated with concise, standard, and fine-grained descriptions, averaging 9.2, 16.5, and 43.6 words, respectively.
\end{itemize}

The high-quality samples excluded from the benchmark constitute \textbf{MRBench-Train}, a training corpus of \textbf{40,168} motion-text pairs drawn from MoCap, in-the-wild, and synthetic sources, which follows the same annotation protocol but remains strictly exclusive from the evaluation set.

\vspace{-3pt}
\section{Granularity-Aware Retrieval Model}
\label{sec:method}
\begin{figure*}[t]
    \centering
    \includegraphics[width=\textwidth]{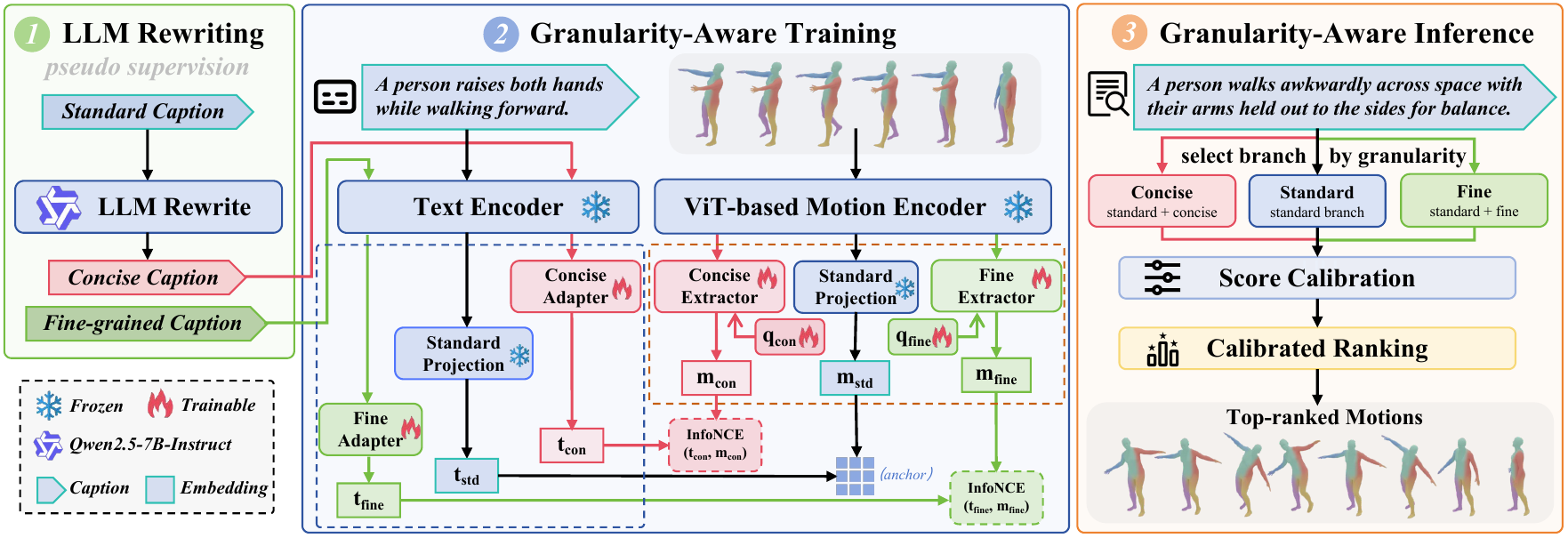}
    \setlength{\abovecaptionskip}{-8pt}
    \caption{\textbf{Overview of our granularity-aware retrieval framework.}
    Standard captions are rewritten into concise and fine-grained variants for pseudo-supervision. A frozen standard-aligned dual encoder provides the global branch, while lightweight adapters and motion extractors learn granularity-specific representations. At inference, the known text granularity selects the calibrated scoring branch. The figure shows text-to-motion retrieval; motion-to-text is symmetric.}
    \label{fig:method}
    \vspace{-2pt}
\end{figure*}

Given $\mathcal{D}_{\mathrm{train}} =\{(m_i,t_i)\}_{i=1}^{N}$ with $N$ motion-text pairs as the training set, we aim to establish a robust cross-modal mapping that aligns motion with multi-granular texts, \ieno, concise, standard, and fine-grained, denoted by $g\in\mathcal{G} =\{\mathrm{con},\mathrm{std},\mathrm{fine}\}$. We perform text-to-motion retrieval to rank candidate motions for a text query $t^g$ and motion-to-text retrieval to rank candidate texts for a motion query $m$.


The predominant motion-text retrieval paradigm considers a dual-encoder model consisting of a motion encoder $E_m$, a text encoder $E_t$, and global projection heads $P_m$ and $P_t$. The global motion/text representations are formulated as 
\begin{equation} 
u_m^{\mathrm{glob}}=P_m(E_m(m)), \quad u_t^{\mathrm{glob}}=P_t(E_t(t)), 
\end{equation} 
Anchored on the base model trained exclusively on a single granularity, we aim to generalize to other textual granularities without compromising its inherent retrieval proficiency.

\subsection{Granularity-Aware Representation}
\label{sec:granularity_representation}
\vspace{-2pt}
We freeze the encoders and global projection heads trained on single granularity alignment as a \emph{global alignment anchor}. Lightweight granularity-specific modules are then introduced for concise and fine-grained descriptions. This design preserves the original retrieval space while learning complementary representations for non-standard queries.

\noindent\textbf{Text-Granularity-Aware Motion Extractors.}
Let $Z_m=[z_1,\ldots,z_T]\in\mathbb{R}^{T\times D}$ denote the token sequence produced by the frozen motion encoder. For each non-standard granularity $g\in\{\mathrm{con},\mathrm{fine}\}$, a learnable query $q_g$ performs attention pooling over the motion tokens as
\begin{equation}
    a_g
    =
    \operatorname{softmax}
    \left(
        \frac{Z_mW_kq_g}{\sqrt{D}}
    \right),
    u_m^g
    =
    P_m^g
    \left(
        a_g^{\top}Z_mW_v
    \right),
\label{eq:granularity_extractor}
\end{equation}
where $W_k$ and $W_v$ are learnable key and value projections, and $P_m^g$ maps the pooled representation into the shared retrieval space. We adopt separate extractors for concise and fine-grained branches, respectively.

\noindent\textbf{Text Granularity Adapters.}
The text representations can be mapped by a granularity-specific projection adapter as
\begin{equation}
    u_t^g=P_t^g(E_t(t^g)),
    \quad
    g\in\{\mathrm{con},\mathrm{fine}\}.
    \label{eq:text_adapter}
\end{equation}
Each adapter $P_t^g$ follows the same architecture of the global text projection and is initialized from $P_t$. We thus reformulate the corresponding adapted similarity as
\begin{equation}
    s_g^{\mathrm{adapt}}(t^g,m)
    =
    \cos\left(u_t^g,u_m^g\right).
    \label{eq:adapted_similarity}
\end{equation}

\subsection{Training and Inference}
\label{sec:training_inference}
\vspace{-2pt}
\noindent\textbf{Two-stage Training.}
We first train the base dual encoder on the standard-caption pairs using a symmetric InfoNCE objective, following standard contrastive representation learning~\cite{oord2018representation,CLIP}. The motion encoder, text encoder, global projection heads, and contrastive temperature are then frozen.

To endow the model with multi-granular alignment capabilities, we expand the text annotations of the training set with concise and fine-grained variants by Qwen2.5-7B-Instruct~\cite{qwen2025qwen25technicalreport}. However, unlike our carefully curated dataset, this text-only expansion leverages the LLM to generate relatively coarse descriptions, which serve strictly as simple pseudo-labels for granularity-specific supervision.
For the second stage, we only optimize the motion extractors and text adapters as 
\begin{equation} \mathcal{L}_{\mathrm{adapt}} = \sum_{g\in\{\mathrm{con},\mathrm{fine}\}} \mathcal{L}_{\mathrm{NCE}} \left(u_t^g,u_m^g\right), \label{eq:adaptation_loss} 
\end{equation}

\begin{table*}[t]
\centering
\small
\setlength{\tabcolsep}{1.5pt}
\begin{tabular}{@{}llc|ccccccc|ccccccc@{}}
\toprule
\multirow{2}{*}{Test}
& \multirow{2}{*}{Text}
& \multirow{2}{*}{Method}
& \multicolumn{7}{c|}{\textbf{Text-to-Motion Retrieval}}
& \multicolumn{7}{c}{\textbf{Motion-to-Text Retrieval}} \\
\cmidrule(lr){4-10}\cmidrule(lr){11-17}
& &
& R@1$\uparrow$
& R@5$\uparrow$
& R@10$\uparrow$
& MedR$\downarrow$
& N@1$\uparrow$
& N@5$\uparrow$
& N@10$\uparrow$
& R@1$\uparrow$
& R@5$\uparrow$
& R@10$\uparrow$
& MedR$\downarrow$
& N@1$\uparrow$
& N@5$\uparrow$
& N@10$\uparrow$ \\
\midrule

\multirow{4}{*}{\rotatebox[origin=c]{90}{H3D}}
& \multirow{4}{*}{Standard}
& TEMOS
& 4.38 & 12.39 & 18.73 & 86.00 & 4.38 & 7.31 & 9.43
& 4.97 & 12.61 & 17.95 & 104.00 & 4.97 & 8.64 & 10.82 \\

& & TMR
& 8.90 & 22.81 & 32.69 & 24.00 & 8.90 & 14.89 & 18.63
& 9.22 & 22.79 & 32.48 & 26.00 & 9.22 & 15.62 & 19.59 \\

& & MoPatch
& \underline{10.80}
& \underline{26.72}
& \underline{38.02}
& \underline{19.00}
& \underline{10.80}
& \underline{19.01}
& \underline{23.57}
& \underline{11.25}
& \underline{26.86}
& \underline{37.40}
& \underline{20.50}
& \underline{11.25}
& \underline{19.98}
& \underline{24.33} \\

& & SGAR
& \textbf{12.86}
& \textbf{30.75}
& \textbf{43.00}
& \textbf{15.00}
& \textbf{12.86}
& \textbf{20.84}
& \textbf{25.67}
& \textbf{13.82}
& \textbf{30.09}
& \textbf{41.83}
& \textbf{16.00}
& \textbf{13.82}
& \textbf{21.82}
& \textbf{26.23} \\

\midrule

\multirow{15}{*}{\rotatebox[origin=c]{90}{MRBench}}

& \multirow{5}{*}{Concise}
& TEMOS
& 0.86 & 3.45 & 5.78 & 498.00 & 0.86 & 1.88 & 2.56
& 0.56 & 2.18 & 3.81 & 588.00 & 0.56 & 1.38 & 1.91 \\

& & TMR
& 2.18 & 8.67 & 13.36 & 230.50 & 2.12 & 5.00 & 6.45
& 1.47 & 5.63 & 9.59 & 294.00 & 1.47 & 3.55 & 4.88 \\

& & MoPatch
& \underline{5.60}
& \textbf{18.05}
& \textbf{24.99}
& 82.00
& 5.43
& \textbf{10.92}
& \underline{13.27}
& 3.42
& \textbf{11.80}
& \textbf{18.02}
& \underline{147.00}
& 3.42
& \underline{7.67}
& \underline{9.75} \\

& & SGAR
& \textbf{5.63}
& 17.64
& \underline{24.90}
& \textbf{73.50}
& \textbf{5.66}
& \underline{10.86}
& \textbf{13.34}
& \textbf{3.72}
& \underline{11.24}
& 16.58
& 165.00
& \textbf{3.72}
& 7.63
& 9.38 \\

& & Ours
& \underline{5.60}
& \underline{17.70}
& \underline{24.90}
& \underline{79.00}
& \underline{5.60}
& 10.85
& \underline{13.27}
& \underline{3.57}
& \textbf{11.80}
& \underline{17.99}
& \textbf{132.00}
& \underline{3.57}
& \textbf{7.77}
& \textbf{9.87} \\

\cmidrule(lr){2-17}

& \multirow{5}{*}{Standard}
& TEMOS
& 1.47 & 5.16 & 8.08 & 424.00 & 1.47 & 3.38 & 4.32
& 1.18 & 3.57 & 5.49 & 496.00 & 1.18 & 2.37 & 2.99 \\

& & TMR
& 4.13 & 11.53 & 17.67 & 181.50 & 4.16 & 7.95 & 9.91
& 2.33 & 8.85 & 12.98 & 217.00 & 2.33 & 5.70 & 7.01 \\

& & MoPatch
& \textbf{7.14}
& \textbf{20.12}
& \textbf{27.08}
& \underline{79.00}
& \textbf{7.11}
& \textbf{13.85}
& \textbf{16.10}
& \textbf{5.66}
& \textbf{15.13}
& \textbf{21.27}
& \textbf{118.00}
& \textbf{5.66}
& \textbf{10.47}
& \textbf{12.46} \\

& & SGAR
& \underline{6.31}
& \underline{18.67}
& \underline{25.87}
& \textbf{73.50}
& \underline{6.46}
& \underline{12.83}
& \underline{15.16}
& \underline{4.48}
& \underline{12.89}
& \underline{18.85}
& \underline{134.00}
& \underline{4.48}
& \underline{8.79}
& \underline{10.68} \\

& & Ours
& \textbf{7.14}
& \textbf{20.12}
& \textbf{27.08}
& \underline{79.00}
& \textbf{7.11}
& \textbf{13.85}
& \textbf{16.10}
& \textbf{5.66}
& \textbf{15.13}
& \textbf{21.27}
& \textbf{118.00}
& \textbf{5.66}
& \textbf{10.47}
& \textbf{12.46} \\

\cmidrule(lr){2-17}

& \multirow{5}{*}{Fine-grained}
& TEMOS
& 0.74 & 2.80 & 4.54 & 468.50 & 0.74 & 1.75 & 2.29
& 0.59 & 2.18 & 3.95 & 478.00 & 0.59 & 1.41 & 1.97 \\

& & TMR
& 1.53 & 6.40 & 10.29 & 188.00 & 1.59 & 3.98 & 5.24
& 1.56 & 5.69 & 9.12 & 227.00 & 1.56 & 3.64 & 4.72 \\

& & MoPatch
& 4.40
& 14.04
& 20.47
& 96.00
& 4.37
& 9.34
& 11.42
& \underline{3.24}
& \underline{10.74}
& \underline{16.73}
& \underline{132.00}
& \underline{3.24}
& \underline{7.01}
& \underline{8.95} \\

& & SGAR
& \underline{4.57}
& \underline{14.63}
& \underline{22.54}
& \underline{76.00}
& \underline{4.57}
& \underline{9.64}
& \underline{12.15}
& 3.19
& 10.71
& 16.70
& 135.00
& 3.19
& 6.94
& 8.86 \\

& & Ours
& \textbf{5.10}
& \textbf{15.84}
& \textbf{22.60}
& \textbf{72.00}
& \textbf{5.01}
& \textbf{10.54}
& \textbf{12.72}
& \textbf{3.92}
& \textbf{11.59}
& \textbf{17.70}
& \textbf{111.50}
& \textbf{3.92}
& \textbf{7.82}
& \textbf{9.79} \\

\bottomrule
\end{tabular}
\setlength{\abovecaptionskip}{3pt}
\caption{Full-set retrieval results of models trained on H3D and evaluated on H3D and MRBench. All texts and motions in the test set form a single candidate pool. N@K denotes nDCG@K. 
The best and second-best results under each test setting and text granularity are highlighted in bold and underlined, respectively.}
\label{tab:hml3d_to_mrbench_fullset}
\vspace{-2pt}
\end{table*}

\noindent\textbf{Granularity-Aware Inference.}
At inference time, the text granularity is dictated by the evaluation
protocol and then guides the activation of the corresponding representation branch. The final score for a query-motion pair $(t^g,m)$ is
\begin{equation}
s(t^g,m)=
\begin{cases}
s_{\mathrm{global}}(t^g,m),
& g=\mathrm{std},\\
\hat{s}_g(t^g,m),
& g\in\{\mathrm{con},\mathrm{fine}\},
\end{cases}
\label{eq:granularity_score}
\end{equation}
where the score for a non-standard granularity is
\begin{equation}
\hat{s}_g(t^g,m)=
\frac{
s_{\mathrm{global}}(t^g,m)
+\alpha_g s_g^{\mathrm{adapt}}(t^g,m)
}{
1+\alpha_g
}.
\label{eq:adapted_score_fusion}
\end{equation}
The fusion weight $\alpha_g$ is selected on the validation split and fixed during inference.

Both terms in~\cref{eq:adapted_score_fusion} are computed from the same input text by the standard and granularity-specific branches, requiring no paired standard caption at inference. The fusion weights are fixed before evaluation and are not tuned on the test set. Text-to-motion retrieval uses the branch corresponding to the query granularity, while motion-to-text retrieval applies the corresponding branch to each candidate description. In the mixed-granularity protocol, the normalization makes fused scores across different textual granularities directly comparable. The holistic framework is shown in~\cref{fig:method}.

\section{Experiments}
\subsection{Experimental Setup}
\vspace{-2pt}
\noindent\textbf{Datasets and Baselines.}
We conduct experiments on HumanML3D~(H3D), KIT-ML, MotionMillion~\cite{humanml3d,plappertKITMotionLanguage2016,fanGoZeroMotionMillion2025}, and MRBench. MRBench is the primary evaluation benchmark and provides concise, standard, and fine-grained descriptions for 3,390 motions. We additionally use MRBench-Train, a training set of 40,168 standard-caption motion-text pairs that is strictly disjoint from the MRBench test set. We evaluate TEMOS~\cite{petrovich22temos}, TMR~\cite{petrovichTMRTexttoMotionRetrieval2023}, MoPatch~\cite{yu2024exploring}, and SGAR~\cite{zhangsgar2024} as baselines. Our granularity-aware model is built upon MoPatch.
Implementation details are provided in Suppl.

\noindent\textbf{Evaluation Protocols and Metrics.}
Unless otherwise specified, we conduct full-set bidirectional retrieval under the concise, standard, fine-grained, and Mixed3 settings, reporting Recall@$K$, nDCG@$K$ ($K\in\{1,5,10\}$), and MedR. In Mixed3 motion-to-text retrieval, all three descriptions associated with a motion are considered relevant; further protocol details are provided in the supplementary material.

\subsection{Main Results}
\vspace{-3pt}
\noindent\textbf{Comprehensive Evaluation on MRBench.}
\cref{tab:hml3d_to_mrbench_fullset} reveals a substantial generalization gap between H3D and MRBench. Models that perform well on the homogeneous H3D test set degrade consistently on MRBench, indicating that conventional \textbf{in-domain evaluation overestimates robustness to broader motion sources and semantic categories}. By incorporating heterogeneous motion distributions and three levels of textual granularity, MRBench exposes both cross-distribution and cross-granularity weaknesses that are not observable on existing benchmarks.

Across methods, standard descriptions remain the easiest, whereas concise and fine-grained queries lead to clear performance drops, showing that \textbf{current retrieval models are sensitive to linguistic granularity}. Our method preserves the standard-caption alignment of the MoPatch anchor while improving fine-grained retrieval in both directions, validating the effectiveness of granularity-specific adaptation.

\begin{table*}[t]
\centering
\small
\setlength{\tabcolsep}{1.5pt}
\begin{tabular}{@{}ll|ccccccc|ccccccc@{}}
\toprule
\multirow{2}{*}{Trainset} & \multirow{2}{*}{Method}
& \multicolumn{7}{c|}{\textbf{Text-to-Motion Retrieval}}
& \multicolumn{7}{c}{\textbf{Motion-to-Text Retrieval}} \\
\cmidrule(lr){3-9}\cmidrule(lr){10-16}
&
& R@1$\uparrow$ & R@5$\uparrow$ & R@10$\uparrow$
& MedR$\downarrow$
& N@1$\uparrow$ & N@5$\uparrow$ & N@10$\uparrow$
& R@1$\uparrow$ & R@5$\uparrow$ & R@10$\uparrow$
& MedR$\downarrow$
& N@1$\uparrow$ & N@5$\uparrow$ & N@10$\uparrow$ \\
\midrule

\multirow{5}{*}{KIT-ML}
& TEMOS
& 0.12 & 0.60 & 1.18 & 1317.50 & 0.12 & 0.36 & 0.53
& 0.09 & 0.56 & 0.94 & 1926.00 & 0.09 & 0.17 & 0.23 \\

& TMR
& 0.29 & 1.20 & 2.15 & 1039.00 & 0.29 & 0.68 & 0.97
& 0.41 & 0.91 & 1.74 & 1476.00 & 0.41 & 0.35 & 0.54 \\

& MoPatch
& 0.65 & 2.44 & 4.35 & 719.00 & 0.65 & 1.41 & 1.96
& \underline{0.59} & 2.27 & \underline{3.95} & 925.00
& \underline{0.59} & \underline{0.82} & \underline{1.15} \\

& SGAR
& \textbf{1.10} & \textbf{4.57} & \textbf{7.80}
& \textbf{379.00}
& \textbf{1.11} & \textbf{2.62} & \textbf{3.65}
& \textbf{0.86} & \textbf{3.42} & \textbf{5.31}
& \textbf{568.00}
& \textbf{0.86} & \textbf{1.14} & \textbf{1.58} \\

& Ours
& \underline{0.68} & \underline{2.78} & \underline{4.77}
& \underline{630.00}
& \underline{0.68} & \underline{1.61} & \underline{2.23}
& 0.53 & \underline{2.30} & 3.69 & \underline{848.00}
& 0.53 & 0.71 & 1.00 \\

\midrule

\multirow{5}{*}{H3D}
& TEMOS
& 1.07 & 3.82 & 6.14 & 463.00 & 1.06 & 2.34 & 3.07
& 1.06 & 3.13 & 5.52 & 537.00 & 1.06 & 1.07 & 1.51 \\

& TMR
& 2.63 & 8.96 & 13.80 & 200.00 & 2.63 & 5.65 & 7.20
& 1.98 & 7.79 & 11.83 & 256.00 & 1.98 & 2.64 & 3.52 \\

& MoPatch
& \underline{5.72} & \underline{17.41} & 24.20 & 86.00
& \underline{5.64} & \underline{11.37} & \underline{13.60}
& \underline{5.07} & \underline{13.10} & \underline{19.73}
& \underline{142.00}
& \underline{5.07} & \underline{5.34} & \underline{7.04} \\

& SGAR
& 5.53 & 17.01 & \underline{24.46} & \textbf{75.00}
& 5.59 & 11.12 & 13.55
& 4.13 & 12.04 & 16.99 & 196.50 & 4.13 & 5.01 & 6.43 \\

& Ours
& \textbf{5.96} & \textbf{17.91} & \textbf{24.88}
& \underline{76.00}
& \textbf{5.92} & \textbf{11.75} & \textbf{14.03}
& \textbf{5.34} & \textbf{14.22} & \textbf{20.21}
& \textbf{131.50}
& \textbf{5.34} & \textbf{5.53} & \textbf{7.20} \\

\midrule

\multirow{4}{*}{MotionMillion}
& TEMOS
& 1.36 & 5.24 & 8.83 & 263.50 & 1.34 & 3.12 & 4.25
& 2.30 & 7.73 & 11.86 & 208.00
& 2.30 & 2.83 & 3.68 \\

& TMR
& \underline{4.35} & \underline{15.07} & \underline{23.79}
& \underline{51.00}
& \underline{4.32} & \underline{9.47} & \underline{12.29}
& \textbf{5.46} & \underline{15.87} & \underline{23.95}
& \textbf{60.00}
& \textbf{5.46} & \textbf{6.10} & \textbf{8.11} \\

& MoPatch
& 3.50 & 11.39 & 17.32
& 111.00
& 3.57 & 7.33 & 9.20
& 2.51 & 8.08 & 12.45
& 258.50
& 2.51 & 3.01 & 4.15 \\

& Ours
& \textbf{4.81} & \textbf{16.33} & \textbf{25.21}
& \textbf{45.00}
& \textbf{4.84} & \textbf{10.31} & \textbf{13.19}
& \underline{4.99} & \textbf{17.05} & \textbf{24.19}
& \underline{61.50}
& \underline{4.99} & \underline{5.12} & \underline{6.17} \\
\midrule

\multirow{4}{*}{\shortstack{MRBench-\\Train}}
& TEMOS
& 3.62 & 10.93 & 16.17 & 164.50 & 3.63 & 7.03 & 8.69
& 6.43 & 17.23 & 23.66 & 75.00 & 6.43 & 6.31 & 7.70 \\

& TMR
& 9.64 & 23.61 & 32.39 & 34.00 & 9.63 & 16.54 & 19.41
& 16.28 & 34.40 & 44.81 & 14.00
& 16.28 & 14.36 & 17.08 \\

& MoPatch
& \underline{11.75} & \underline{28.95} & \underline{38.56}
& \underline{21.00}
& \underline{11.60} & \underline{20.06} & \underline{23.20}
& \underline{17.05} & \underline{37.73} & \underline{47.91}
& \textbf{12.00}
& \underline{17.05} & \underline{16.52} & \underline{19.63} \\

& Ours
& \textbf{12.82} & \textbf{32.04} & \textbf{43.13}
& \textbf{16.00}
& \textbf{12.77} & \textbf{22.28} & \textbf{25.91}
& \textbf{17.91} & \textbf{37.82} & \textbf{48.02}
& \textbf{12.00}
& \textbf{17.91} & \textbf{16.74} & \textbf{20.26} \\

\bottomrule
\end{tabular}
\setlength{\abovecaptionskip}{3pt}
\caption{Cross-dataset full-set retrieval comparison on MRBench under Mixed3, where concise, standard, and fine-grained descriptions jointly form a single candidate pool. N@K denotes nDCG@K. The best and second-best results within each training-set group are highlighted in bold and underlined, respectively.}
\label{tab:cross_dataset_mixed3_fullset}
\end{table*}

\noindent\textbf{Unified Cross-Dataset Evaluation.}
\cref{tab:cross_dataset_mixed3_fullset} shows that MRBench supports a unified comparison of models trained on heterogeneous motion-language corpora. The large performance variation across training sets reveals substantial distributional sensitivity: increasing data scale alone does not guarantee stronger transfer, whereas broader motion coverage and retrieval-oriented annotations are more beneficial for robust motion-text alignment. This further indicates that cross-dataset generalization depends more on the diversity and discriminability of the training distribution.
Models trained on MRBench-Train achieve the strongest overall performance, and our method further improves the MoPatch anchor under the challenging Mixed3 protocol. These results establish MRBench as a general-purpose testbed for evaluating both cross-dataset transfer and mixed-granularity retrieval.

\vspace{-4pt}
\subsection{Ablation Study}
\vspace{-2pt}
\cref{tab:ablation} confirms the motion extractor and text adapter provide complementary benefits for fine-grained alignment. Removing either component generally reduces fine-grained performance, although their effects differ across retrieval directions. Meanwhile, the single-query extractor is more effective than higher-capacity or token-wise alternatives. Score calibration is particularly important for Mixed3: removing it preserves single-granularity performance but substantially degrades motion-to-text retrieval, indicating a mismatch between the score distributions of different granularity branches. Overall, the complete model provides the best balance across concise, fine-grained, and mixed-granularity retrieval while preserving the frozen standard alignment.

\begin{table}[t]
\centering
\small
\setlength{\tabcolsep}{3pt}
\begin{tabular}{@{}l|ccc@{}}
\toprule
Method & Concise & Fine-grained & Mixed3 \\
\midrule
Base model
& 11.30 / 11.74
& 6.84 / 10.03
& 11.75 / 17.05 \\
\midrule
w/o motion extractor
& \underline{11.42} / \underline{11.92}
& 9.14 / \underline{10.06}
& \underline{12.56} / \textbf{18.14} \\
w/o text adapter
& 10.97 / 11.50
& 8.91 / 9.73
& 12.32 / \underline{18.05} \\
w/o score calibration
& \textbf{11.59} / \textbf{12.01}
& \textbf{10.06} / \textbf{10.29}
& \textbf{12.82} / 13.39 \\
4-query extractor
& 11.06 / 11.68
& \underline{9.59} / 9.73
& 12.35 / 17.85 \\
token-wise
& 11.21 / 11.36
& 8.20 / 9.38
& 12.10 / 17.91 \\
\midrule
Ours
& \textbf{11.59} / \textbf{12.01}
& \textbf{10.06} / \textbf{10.29}
& \textbf{12.82} / 17.91 \\
\bottomrule
\end{tabular}
\setlength{\abovecaptionskip}{3pt}
\caption{Ablation results on MRBench under the full-set protocol. Each entry reports R@1 (\%) for text-to-motion / motion-to-text retrieval. Standard results are omitted because the frozen anchor remains unchanged across all variants. The best and second-best distinct results are highlighted in bold and underlined, respectively.}
\vspace{-3pt}
\label{tab:ablation}

\end{table}

\section{Related Works}
\noindent\textbf{Motion-Text Dataset.}
Language annotations in human motion datasets have progressed from predefined action labels~\cite{guo2020action2motion,punnakkalBABELBodiesAction2021,AMASS:ICCV:2019} to free-form sequence-level captions for open-vocabulary generation and retrieval~\cite{plappertKITMotionLanguage2016,humanml3d,hwang2026snapmogen,luHumanTOMATOTextalignedWholebody2023}. More recent datasets further provide body-part-aware, temporally grounded, and expressive descriptions that capture finer motion details~\cite{li2024motion,zhang2023finemogen,zhangsgar2024,hwang2026snapmogen,vimogen,xuDenseMotionCaptioning2026}.
Nevertheless, existing datasets~\cite{lin2023motion, zhang2025motion} generally treat motion-text correspondence at a single, undifferentiated semantic level: multiple captions may be provided, but they are not explicitly organized by granularity, leaving their hierarchical relationships unexplored.

\noindent\textbf{Motion-Text Retrieval.}
Motion-text retrieval ranks motion sequences by their semantic relevance to a textual query. Early methods learn global motion-text embeddings and use retrieval primarily to evaluate cross-modal alignment~\cite{humanml3d,tevetMotionCLIPExposingHuman2022}. Subsequent works treat retrieval as an independent task, improving alignment through contrastive and generative objectives, negative filtering, data augmentation, and cross-dataset training~\cite{petrovichTMRTexttoMotionRetrieval2023,bensabath2024cross,liLaMPLanguageMotionPretraining2025}. Recent methods further introduce spatially structured or body-part features to capture local motion-language correspondences~\cite{yu2024exploring,zhangsgar2024}. 
However, current benchmarks lack the motion diversity and discriminative annotations required to fairly evaluate generalizable motion-text alignment.

\noindent\textbf{Multi-granular Retrieval.}
Prior cross-modal retrieval studies investigate multi-level alignment across visual structures or textual semantics~\cite{gao2022pyramidclip,ging2020coot,jiang2022tencent,chen2020fine}. Recent image-text benchmarks further show that retrieval performance varies with caption specificity, motivating separate evaluation of coarse and fine-grained descriptions~\cite{chen2023rethinking,hendriksen2025benchmark}. However, existing work rarely provides well-organized, unified semantic levels for evaluation. 

\section{Conclusion}
MRBench reframes motion-text retrieval evaluation from motion and text perspectives. We find that strong in-domain retrieval does not guarantee robust transfer, while mixed-granularity ranking requires granularity-aware representations. Our granularity-aware retrieval model further suggests that robustness to non-standard descriptions can be improved without sacrificing established standard-caption alignment. Together, the benchmark and model provide a practical foundation for developing motion-language systems that generalize beyond narrow datasets and fixed caption styles.

\bibliography{aaai2027}

\end{document}